\documentclass[runningheads]{llncs}

\usepackage{eccv}

\usepackage{eccvabbrv}
\usepackage{graphicx}
\usepackage{booktabs}
\usepackage[accsupp]{axessibility}
\usepackage{subcaption}
\usepackage{placeins}
\usepackage[table]{xcolor}
\usepackage{comment}
\usepackage[bookmarks=false]{hyperref}
\usepackage{orcidlink}

\definecolor{bestred}{RGB}{248,205,205}
\definecolor{secondorange}{RGB}{255,218,153}
\definecolor{thirdyellow}{RGB}{255,242,153}

\newcommand{\best}[1]{\cellcolor{bestred}{#1}}
\newcommand{\secondbest}[1]{\cellcolor{secondorange}{#1}}
\newcommand{\thirdbest}[1]{\cellcolor{thirdyellow}{#1}}
\newcommand{\ud}{\mathrm{d}}
\newcommand{\TrainLoss}{\mathcal{L}_{\mathrm{train}}}

\newcommand{\qualw}{0.15\textwidth}

\usepackage{array}

\newcolumntype{S}{>{\centering\arraybackslash}m{0.035\textwidth}}
\newcolumntype{Q}{>{\centering\arraybackslash}m{\qualw}}

\newcommand{\sceneLabel}[1]{%
    \rotatebox[origin=c]{90}{#1}%
}

\newcommand{\qualimg}[1]{%
    \includegraphics[width=\linewidth]{#1}%
}

\begin{document}

% ---------------------------------------------------------------
% Paper metadata
\title{Point-Based 3D Reconstruction from Sparse Views under Known Illumination}

\titlerunning{Point-Based 3D Reconstruction from Sparse Views}

\author{
Magnus Kaufmann Gjerde\inst{1}\orcidlink{0009-0000-2799-9035} \and
Joakim Bruslund Haurum\inst{2,5}\orcidlink{0000-0002-0544-0422} \and
Jeppe Revall Frisvad\inst{3}\orcidlink{0000-0002-0603-3669} \and
Markus Worchel\inst{4}\orcidlink{0000-0002-3469-6750} \and
J. Andreas Bærentzen\inst{3}\orcidlink{0000-0003-2583-0660} \and
Thomas B. Moeslund\inst{1,5}\orcidlink{0000-0001-7584-5209}
}
\authorrunning{M. K. Gjerde et al.}

\institute{
Visual Analysis \& Perception Lab, Aalborg University
\and
Centre for Software Technology, University of Southern Denmark
\and
Department of Applied Mathematics and Computer Science, Technical University of Denmark
\and
Technische Universität Berlin
\and
Pioneer Centre for Artificial Intelligence
}

\maketitle

% ---------------------------------------------------------------
\begin{abstract}
Sparse view 3D reconstruction is commonly addressed with neural implicit surfaces or dense point-based representations such as Gaussian splatting. Surface-aware splatting methods improve extracted geometry through oriented primitives and regularization, while RadiosityGS incorporates differentiable light transport through a radiosity inspired finite-element surfel formulation. We propose a differentiable point rendering method based on opacity-bearing beta surfels. An opacity explicit adjoint light transport formulation provides gradients for surfel geometry and appearance parameters, allowing physically based light transport to constrain reconstruction.
Across five synthetic objects reconstructed from ten posed views, our method achieves the lowest mean symmetric Chamfer distance among the evaluated baselines and reduces mean Chamfer distance by 28.5\% relative to the strongest point-based baseline while using only 267 surfels on average, approximately 161× fewer primitives. Directional Chamfer results further show improved accuracy and competitive completion relative to related point-based methods. These results show that, in the controlled direct illumination setting, compact beta surfels combined with transport-based optimization can recover surfaces without relying on the tens to hundreds of thousands of primitives used by the evaluated baselines.
\end{abstract}

% --------------------------------------------------------------
\section{Introduction}
\label{sec:introduction}

Recovering accurate geometry from a small set of images remains challenging because appearance alone does not uniquely determine surface shape, reflectance, and visibility. Recent point-based methods, such as Gaussian Splatting (GS)~\cite{kerbl20233d}, can synthesize high-quality images by optimizing large sets of primitives. However, image fidelity does not necessarily imply accurate geometry: Optimized primitives may compensate for photometric error through opacity and their appearance parameters rather than converging to a coherent surface. Consequently, meshes extracted from appearance-driven splatting representations can contain floaters, fragmented regions, and excessive geometric detail. Surface-aware variants such as 2DGS~\cite{huang20242d} and SuGaR~\cite{guedon2023sugar} improve geometric behavior through oriented primitives and regularization, but they still rely on rasterization-based image formation rather than physically based light transport.

Known illumination provides an additional source of geometric supervision. Under calibrated point lighting and Lambertian reflectance, surface shape and orientation affect observed radiance through shading and visibility. Physics-based differentiable rendering provides a principled way to exploit these effects by differentiating image formation through physical light transport~\cite{Li:2018:DMC,Zhang:2019:DTRT}.

RadiosityGS~\cite{jiang2025differentiable} combines Gaussian surfels with a radiosity-inspired finite-element formulation of differentiable light transport for relighting and geometry reconstruction. Its formulation represents radiometric quantities using per-surfel coefficients and assumes spatially constant material, emission, and outgoing radiance within each Gaussian-surfel support. This enables efficient view-independent transport computation and motivates our alternative formulation, which, in contrast, expresses transport through surface-level quantities evaluated at the specific local coordinates of each surfel interaction, allowing radiance and material response to vary across the surfel support.

\begin{figure}[!t]
\centering
\includegraphics[width=\linewidth]{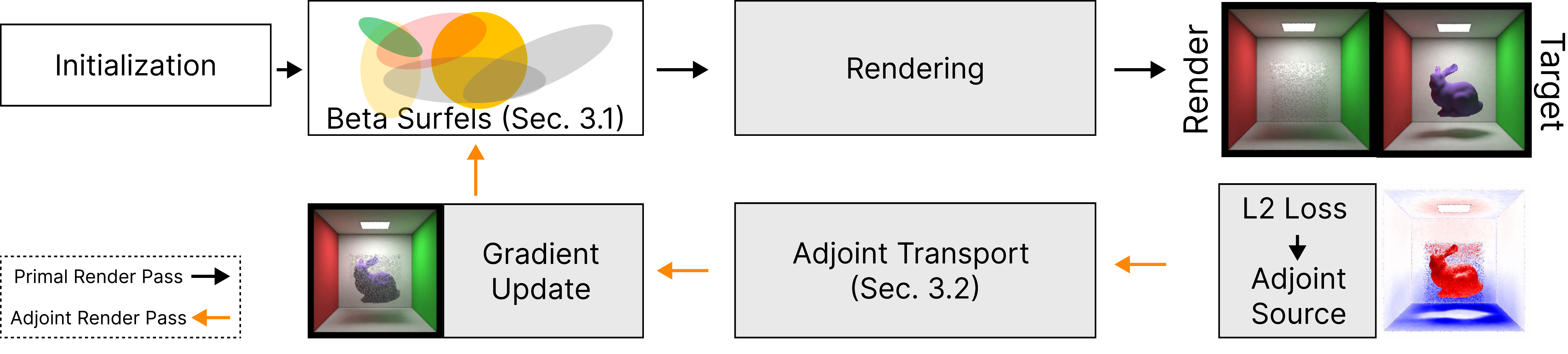}
\caption{
Overview of our optimization pipeline.
Starting from a sparse beta-surfel initialization, we alternate primal rendering, residual evaluation, adjoint transport, and parameter updates.
The adjoint source is the image-space residual between the rendered and reference images.
}
\label{fig:system_overview}
\end{figure}
We adopt this local-coordinate view for compact surface recovery. Our representation uses opacity-bearing beta surfels~\cite{liu2025deformablebetasplatting} with spatially varying local support. Rather than assigning one outgoing radiance value to an entire primitive, we evaluate scattering, transmittance, and visibility at sampled surfel coordinates. We then specialize continuous adjoint light transport to this representation, retaining segment transmittance explicitly to obtain analytical parameter gradients for surfel geometry, material, opacity, and local opacity-profile shape. We estimate the resulting transport integrals with Monte Carlo path sampling and evaluate the direct illumination specialization under calibrated point lights.

Figure~\ref{fig:system_overview} summarizes the optimization loop. Starting from a sparse beta-surfel initialization, primal rendering produces image residuals that define the adjoint source. The adjoint transport pass then provides gradients used to update surfel geometry, material, opacity, and kernel shape.

We evaluate the method on five synthetic objects reconstructed from ten posed input views under calibrated static near-field point illumination. Our method obtains the lowest mean symmetric Chamfer distance among the evaluated baselines while using only 267 surfels on average. In comparison, the standard 2DGS 30K configuration uses 51k primitives on average. Directional Chamfer measurements show the lowest reconstruction-to-ground-truth error and competitive ground-truth-to-reconstruction error. Although we focus on isolated objects rather than large-scale scenes, we address object-centric 3D reconstruction by recovering geometry with orders of magnitude fewer primitives than the evaluated point-based baselines while maintaining better reconstruction results.

Our contributions are:
\begin{itemize}
\item A method for sparse-view reconstruction under known direct illumination using compact, opacity-bearing beta surfels, with explicit scattering, visibility, and transmittance and without assuming constant outgoing radiance over each surfel.

\item An opacity-explicit adjoint light-transport formulation that provides gradients with respect to surfel geometry, albedo, opacity, and local opacity-profile shape.
\end{itemize}

\section{Related Work}
\label{sec:related_work}

\paragraph{Physics-based differentiable rendering.}
This approach optimizes scene parameters by differentiating image formation with respect to geometry, materials, or illumination. Early Monte Carlo methods developed estimators for visibility changes and transport derivatives under complex light transport~\cite{Li:2018:DMC,Zhang:2020:PSDR,Zhang:2019:DTRT}. Adjoint formulations provide an alternative perspective in which gradients are derived at the level of the continuous transport operator before discretization~\cite{nimier2020radiative,stam2020computing}. Recent work has extended differentiable light transport to point-based scene representations. In particular, RadiosityGS~\cite{jiang2025differentiable} formulates transport over Gaussian surfels using a radiosity-inspired finite-element approximation. In contrast, our method instead evaluates transport using generic Monte Carlo path tracing, without a constant local radiance assumption.

\paragraph{Surface reconstruction from posed images.}
Neural implicit methods such as NeuS~\cite{wang2023neuslearningneuralimplicit} and NeuS2~\cite{neus2} optimize continuous implicit surfaces from posed images, while GeoSVR~\cite{li2025geosvr} uses sparse voxels and geometry-oriented regularization. These methods contextualize our explicit surfel representation, which directly exploits calibrated point-light transport.

\paragraph{Point-based reconstruction and Gaussian splatting.}
Point-based rendering has a long history in graphics, and Gaussian splatting methods have recently enabled high-quality novel-view synthesis with optimized primitive sets~\cite{zwicker2002ewa,kerbl20233d}. Surface-aware variants improve geometric behavior by using oriented primitives or additional regularization. In particular, 2DGS~\cite{huang20242d} represents scenes using oriented 2D Gaussian disks and introduces depth-distortion and normal-consistency losses for mesh reconstruction. SuGaR~\cite{guedon2023sugar} aligns Gaussian primitives with an extracted surface to improve mesh quality. Our representation is similarly surface-oriented, but differs in its compact-support opacity kernel, continuous transmittance model, and physics-based optimization under known illumination.

\section{Method}
\label{sec:method}
Our goal is to recover scene parameters $\psi$ (geometry, materials, etc.), whose rendered measurements match the posed reference images under known direct illumination, while evaluating transport at local surfel coordinates rather than assuming constant radiance over each support. We first discuss how we define our geometry primitive before proceeding to differentiate light transport with respect to the scene parameters.

\subsection{Point Geometry Parameterization}
\label{sec:point_geometry}
We introduce our beta surfel elements by expressing them as regular surfaces. We do so by using the theory of Do Carmo \cite{do2016differential}. A beta surfel represents a finite planar surface element embedded in $\mathbb{R}^3$ and parameterized by a local tangent frame. Each surfel $k$ is defined by a learnable set of parameters:
\begin{equation}\label{eq:surfel_parameters}
    \psi_k = \{ \mathbf{p}_k,\ \mathbf{t}_{u,k},\ \mathbf{t}_{v,k},\ \mathbf{s}_k,\ \mathbf{\rho}_k,\ \eta_k,\ \beta_k \}\,,
\end{equation}
where $\mathbf{p}_k \in \mathbb{R}^3$ is the surfel center, $\mathbf{t}_{u,k}$ and $\mathbf{t}_{v,k}$ are orthonormal tangent directions, defining the surfel normal by $\mathbf{n}_k = \mathbf{t}_{u,k} \times \mathbf{t}_{v,k}$, $\mathbf{s}_k = (s_{u,k}, s_{v,k})$ are scale parameters for the tangent directions, $\mathbf{\rho}_k$ is a diffuse albedo, $\eta_k \in [0,1]$ is a uniform opacity factor, and $\beta_k$ controls the shape of the opacity kernel.

We define a local chart of the regular surface:
\begin{equation}
    {\Phi}_k(u,v) = \mathbf{p}_k + s_{u,k}\mathbf{t}_{u,k}u + s_{v,k}\mathbf{t}_{v,k}v \,,
    \label{eq:surfel_chart}
\end{equation}
with $(u,v) \in \mathbb{R}^2$ as the local coordinates of the surfel.
We also define a constant Jacobian determinant which is found by analyzing the surface with its first fundamental form
\begin{equation}
    \ud{A} = s_{u,k} s_{v,k}\,\ud{u}\,\ud{v} \,,
\end{equation}
which we use for reparameterizing the transport operator integral over beta surfels.

We use the local coordinates $(u, v)$ to model continuous visibility and transmittance $\alpha_k^{\mathrm{geom}}$. While Gaussian kernels are commonly used in point-based rendering, we instead use a beta kernel presented by Liu \etal~\cite{liu2025deformablebetasplatting}:
\begin{equation}
\alpha_k^{\mathrm{geom}}(u,v;\beta_k)
=
\begin{cases}
(1-r^2)^{b(\beta_k)}, & r^2 \leq 1,\\
0, & r^2 > 1,
\end{cases}
\end{equation}
with
\begin{equation}
b(\beta_k)=4e^{\beta_k}
\qquad\text{and}\qquad
r^2=u^2+v^2.
\end{equation}
When $\beta_k=0$ the kernel starts with a Gaussian-like function. 
The final effective opacity of the surfel is
\begin{equation}
    \alpha_k(u,v) = \eta_k\,\alpha_k^{\mathrm{geom}}(u,v;\beta_k) \,,
\end{equation}
where $0 \le \alpha_k \le 1$.
Later in ablation studies it is shown that optimizing the beta profile improves reconstruction on our data relative to the fixed Gaussian-like $\beta_k=0$ profile. Furthermore, a benefit of the compactly supported beta kernel is that it removes the need for a heuristic footprint cutoff which can vary between different Gaussian Splatting implementations.

%With the transport-gradient estimator and beta-surfel parameterizationspecified, we now instantiate the optimization for sparse-viewreconstruction under known direct illumination. The following setup uses thederived gradients to jointly optimize surfel geometry, orientation, scale,diffuse albedo, opacity, and local opacity-profile shape from posed referenceimages.

As shown with our definition of opacity, a beta surfel is not an opaque surface element. Its local opacity profile both determines scattering at a sampled point and attenuates every light or camera connection passing through its support. Consequently, changing a surfel parameter affects image formation in two ways, either through local scattering or segment transmittance. Monte Carlo implementations can account for opacity through stochastic path continuation which is sufficient for rendering, but it does not directly expose the derivatives of the continuous transmittance needed to differentiate analytically. We therefore retain a segment transmittance defined as $\tau_\psi$ explicitly before forming the light transport gradients.

This also differs from the primitive-level finite-element discretization used by RadiosityGS~\cite{jiang2025differentiable}, which represents radiometric quantities using per-surfel coefficients and assumes them spatially constant within each Gaussian-surfel support. In contrast, we estimate transport at sampled local coordinates within each beta-surfel support.

\subsection{Light Transport Gradients with Explicit Transmittance}
The gradients considered here could alternatively be obtained by applying reverse-mode automatic differentiation to a sampled rendering computation. While this can be practical at first, differentiating recursive transport becomes increasingly demanding as longer path histories and their intermediate states must be stored. We instead follow the adjoint perspective of Radiative Backpropagation~\cite{stam2020computing, nimier2020radiative}, while retaining explicit transmittance $\tau_\psi$. We subsequently show by example the zeroth-order term of the adjoint Neumann series to show that an appropriate parameterization of the transport operator keeps the sensor sample fixed, thereby avoiding camera-response derivatives induced by a moving surface projection.

We optimize scene parameters $\psi$ for all surfels described in the previous section, to obtain a rendered image matching an observed reference image. We write beam transmittance as $\tau_\psi$ in the rendering equation, where the $\psi$ subscript denotes the dependence on the scene parameters $\psi$.
This makes the transmittance terms involving $\partial \tau_\psi / \partial \psi$ explicit and connects the continuous adjoint formulation to our opacity-bearing point primitives.
Assuming a pinhole camera at the position $\mathbf{x_c}$, the intensity captured in one of its pixels is an integral of incident radiance $L_i$ across points $\mathbf{x_p}$ in the projected pixel area, or equivalently an integral of outgoing radiance $L_o$ from observed surface positions $\mathbf{x_o}$ found in the ray direction $\boldsymbol{\omega} = (\mathbf{x_p} - \mathbf{x_c})/t_p$ with $t_p = \|\mathbf{x_p}-\mathbf{x_c}\|$, such that $\mathbf{x_p} = \mathbf{x_c} + t_p\,\boldsymbol{\omega}$ and $\mathbf{x_o} = \mathbf{x_c} + t_o\,\boldsymbol{\omega}$.
A least squares cost function for our optimization is then
\begin{eqnarray} 
\mathcal{J}(L_o,\psi) & = & \int_{A(\psi)}\!\int_{4\pi} \!\!\!W_c(\mathbf{x},\boldsymbol{\omega})\,J(L_o, \psi)\,\ud{\omega}\,dA \label{eq:cost}\\
J(L_o, \psi) & = & \tfrac{1}{2}\!\left(L_o(\mathbf{x}_p,-\boldsymbol{\omega},\psi)-L_{\mathrm{ref}}(\mathbf{x}_p,-\boldsymbol{\omega})\right)^2 \,.
\end{eqnarray}
where $\mathbf{x}$ is a point in $A(\psi)$, which is the union of $\mathbf{x}_c$ and all surface positions in the scene. The domain $4\pi$ refers to the full solid angle of directions $\boldsymbol{\omega}$, $L_{\mathrm{ref}}$ is radiance sampled in $\mathbf{x}_p$ of the reference image, and $W_c = [\boldsymbol{\omega} \in \Omega]\,\delta(\mathbf{x} - \mathbf{x}_c)$ is the camera response function, defined using an Iverson bracket and a Dirac delta function to only include paths observed at $\mathbf{x}_c$ with $\boldsymbol{\omega}$ in the solid angle $\Omega$ subtended by the camera sensor (the film). The optimal parameters are obtained by solving
\begin{equation} \label{eq:optimal}
\psi^{\star} = {\arg\min}_{\psi}\, \mathcal{J}(L_o,\psi) \quad\text{s.t.}\quad R(L_o,\psi)=0 \,,
\end{equation}
where the constraint function $R$ is based on light transport.

In the original discussion of methods for solving the rendering equation, Kajiya~\cite{kajiya1986rendering} mentioned that one option is to use a linear operator $\mathcal{T}_{\psi}$ to represent the integral in the reflected radiance term.
We write the rendering equation including transmittance as:
\begin{equation} \label{eq:rendering}
L_o = \tau_{\psi} (L_e + \mathcal{T}_{\psi}L_o) \,,
\end{equation}
where $L_e$ is emitted radiance and $\tau_{\psi}$ is beam transmittance induced by the opacity-bearing scene representation. This is our main addition to Stam's formulation \cite{stam2020computing}.
The transport operator $\mathcal{T}_{\psi}$ acts on outgoing radiance and includes propagation and scattering between surface elements.
Writing $\tau_\psi$ explicitly is not required in all presentations of light transport, but it is necessary here because the opacity parameters of the point primitives are optimized and therefore contribute directly to the gradient.
 %($L_i = V L_o$, where $V$ is visibility, which 

Using Lagrange multipliers and the continuous adjoint representation we end up with the following expression (more details in supplemental Appendix A):
\begin{equation}   \label{eq:gradient}
\frac{\partial\mathcal{L}}{\partial\psi} =  \frac{\partial\mathcal{J}}{\partial\psi} + \left\langle{\frac{\partial(\mathcal{T}_{\psi}^{\ast}\tau_{\psi})}{\partial\psi}p, L_o}\right\rangle + \left\langle{p, \frac{\partial{\tau_{\psi}}}{\partial\psi}}L_e\right\rangle \,,
\end{equation}
where $\ast$ denotes the adjoint and $p$ is a Lagrangian multiplier traced as incident directional importance. The appeal of the adjoint formulation is that it eliminates the need to compute the variation of $L_o$ with the parameters $\psi$ explicitly. We only need derivatives of the transport operator kernel for each surfel, yielding favorable scaling properties with respect to the number of optimized parameters.

For a fixed camera and camera response function $W_c$, the measurement operator and the image domain are independent of the scene parameters $\psi$. Consequently, the cost functional $\mathcal{J}$ has no explicit dependence on $\psi$, and the first term of Eq.~\eqref{eq:gradient} vanishes.

\subsection{Transport and Transmittance Derivatives}

As derived in Sec. A.1 of the supplementary material, the Lagrangian multiplier $p$ is given as a Neumann series:
\begin{equation} \nonumber
p = \sum_{n=0}^{\infty} (\mathcal{T}_\psi^\ast\tau_{\psi})^n \frac{\partial{J}}{\partial{L_o}}W_c = \sum_{n=0}^{\infty} (\mathcal{T}_\psi^\ast\tau_{\psi})^n (L_o - L_{\mathrm{ref}})\,W_c \,.
\end{equation}
To clearly show how the gradient connects to propagation of this adjoint variable, we take a special look at the zeroth-order term of this Neumann series, and denote it by
\begin{equation}
\begin{aligned}
p_0
=
\left(\mathcal{T}_{\psi}^{*}\tau_{\psi}\right)^0
\left(L_o-L_{\mathrm{ref}}\right)W_c=
\left(L_o-L_{\mathrm{ref}}\right)W_c .
\end{aligned}
\end{equation}
Using the zeroth-order adjoint approximation $p_0 \approx p$, the gradient of the Lagrangian $\mathcal{L}$ with respect to the scene parameters $\psi$ is then
\begin{eqnarray}
\left.\frac{\partial \mathcal{L}}{\partial \psi}\right\vert_{p\approx p_0}
& =&
\left\langle
\frac{\partial\left(\mathcal{T}_{\psi}^{*}\tau_{\psi}\right)}
{\partial\psi}
p_0,
L_o
\right\rangle
+
\left\langle
p_0,
\frac{\partial\tau_{\psi}}{\partial\psi}L_e
\right\rangle .
\nonumber\\
& = &
\left\langle
\frac{\partial\mathcal{T}_{\psi}^{*}}
{\partial\psi}
\tau_{\psi}
p_0,
L_o
\right\rangle
+
\left\langle
p_0,
\frac{\partial\tau_{\psi}}{\partial\psi}
(L_e + \mathcal{T}_{\psi}L_o)
\right\rangle .
\end{eqnarray}
\begin{comment}
\\ \\
\begin{equation}
\left.\frac{\partial \mathcal{L}}{\partial \psi}\right\vert_{p=p_0}
=
\left\langle
p_0,
\frac{\partial\tau_{\psi}}
{\partial\psi}
\mathcal{T}_{\psi}
L_o
\right\rangle
+
\left\langle
\frac{\partial\mathcal{T}_{\psi}^{*}}
{\partial\psi}
\tau_{\psi}
p_0,
L_o
\right\rangle
+
\left\langle
p_0,
\frac{\partial\tau_{\psi}}{\partial\psi}L_e
\right\rangle .
\end{equation}
\begin{equation}
\left.\frac{\partial \mathcal{L}}{\partial \psi}\right\vert_{p=p_0}
=
\left\langle
\frac{\partial\mathcal{T}_{\psi}^{*}}
{\partial\psi}
\tau_{\psi}
p_0,
L_o
\right\rangle
+
\left\langle
p_0,
\frac{\partial\tau_{\psi}}{\partial\psi}
(L_e + \mathcal{T}_{\psi}L_o)
\right\rangle .
\end{equation}
\end{comment}
%
The first term includes derivatives of transport from the camera to the scene. %, including transmittance.
The second term includes derivatives of the transmittance of directly visible emitters and surfaces. In the following, we describe the computation of the transport and transmittance derivatives.

\subsubsection{Transmittance derivative.}
Here, \(\alpha_i(u_i,v_i)\) is the opacity of a surfel, and \(i\) is the \(i\)-th surfel in the front-to-back composition between \(\mathbf{x}_0\) and \(\mathbf{x}_1\):
\begin{equation}
\tau_{\psi}(\mathbf{x}_0,\mathbf{x}_1)
=
\prod_{i\in\mathcal{H}}
\left(1-\alpha_i(u_i,v_i)\right).
\end{equation}
The transmittance derivative, where \(\mathcal{H}\) is the fixed set of segment intersections assuming the ordering and membership of intersected surfels remain unchanged on the path between \(\mathbf{x}_0\) and \(\mathbf{x}_1\), is
\begin{equation}
\frac{\partial\tau_{\psi}(\mathbf{x}_0,\mathbf{x}_1)}{\partial\psi}
=
-
\sum_{k\in\mathcal{H}}
\frac{\partial\alpha_k(u_k,v_k)}{\partial\psi}
\prod_{\substack{i\in\mathcal{H}\\i\neq k}}
\left(1-\alpha_i(u_i,v_i)\right).
\end{equation}
\subsubsection{Transport derivative.}
The adjoint transport operator admits equivalent hemispherical and
surface-area parameterizations. Here, $\boldsymbol{\theta}$ denotes the direction at which the operator is evaluated. In the hemispherical form, $\boldsymbol{\phi}$ parameterizes the direction from
$\mathbf{x}_0$ and determines the next interaction
$\mathbf{x}_1=r_{\psi}(\mathbf{x}_0,\boldsymbol{\phi})$. The two forms are
\begin{eqnarray}
\mathcal{T}_{\psi}^{*}p_0
&=&
\int_{\mathbb{S}^2}
p_0\!\left(r_{\psi}(\mathbf{x}_0,\boldsymbol{\phi}),-\boldsymbol{\phi}\right)
f_s(\mathbf{x}_0,\boldsymbol{\theta},-\boldsymbol{\phi})
\left|\cos\!\left(\mathbf{N}_{\mathbf{x}_0},\boldsymbol{\phi}\right)\right|
\,\mathrm{d}\omega_{\boldsymbol{\phi}},
\\
\mathcal{T}_{\psi}^{*}p_0
&=&
\int_A
p_0(\mathbf{x}_1,-\boldsymbol{\phi})
f_s(\mathbf{x}_0,\boldsymbol{\theta},-\boldsymbol{\phi})
G(\mathbf{x}_0,\mathbf{x}_1)
\,\mathrm{d}A_{\mathbf{x}_1}.
\end{eqnarray}
For primal rendering, either formulation may be used since both represent the same transport operator. Under differentiation, however, the choice of parameterization becomes important, as discussed by Worchel \etal~\cite{worchel2025radiative}. We use different parameterizations for different path segments: Recursive surface-to-surface transport is evaluated using the surface-area form, which permits reparameterization over fixed local surfel coordinates, while the primary camera-to-scene connection is evaluated using the hemispherical form.
For recursive surface-to-surface transport, we locally denote the
current interaction by \(\mathbf{x}_0\) and the next interaction by
\(\mathbf{x}_1=\Phi_k(u,v)\). Its derivative is
\begin{equation}
\frac{\partial\mathcal{T}_{\psi}^{*}}{\partial\psi}p_0
=
\int_U
\partial_{\psi}
\left[
p_0(\mathbf{x}_1,-\boldsymbol{\phi})
f_s(\mathbf{x}_0,\boldsymbol{\theta},-\boldsymbol{\phi})
G(\mathbf{x}_0,\mathbf{x}_1)
J_{\psi}
\right]
\,\mathrm{d}u\,\mathrm{d}v .
\end{equation}
Here, \(U\) denotes the local surfel-coordinate domain and
\(J_{\psi}=s_us_v\) is the corresponding determinant of the surface-area Jacobian.
For the primary camera-to-scene propagation, we instead differentiate the hemispherical form. Let
\(\mathbf{x}_1=r_{\psi}(\mathbf{x}_c,\boldsymbol{\phi})\), where the camera position \(\mathbf{x}_c\) and sampled
sensor direction \(\boldsymbol{\theta}\) are held fixed:
\begin{equation}
\frac{\partial\mathcal{T}_{\psi}^{*}}{\partial\psi}p_0
=
\int_{\mathbb{S}^2}
\partial_{\psi}
\left[
p_0(\mathbf{x}_1,-\boldsymbol{\phi})
f_s(\mathbf{x}_1,\boldsymbol{\theta},-\boldsymbol{\phi})
\left|\cos(\mathbf{N}_{\mathbf{x}_1},\boldsymbol{\phi})\right|
\right]
\,\mathrm{d}\omega_{\boldsymbol{\phi}}.
\end{equation}
A surface-area parameterization would instead evaluate the adjoint
source at the projection of a parameter-dependent surface point,
\(p_0(\mathbf{X}_{\psi}(u,v),\boldsymbol{\omega}_{\psi}(u,v))\). Differentiating this expression
introduces chain-rule terms through the parameter-dependent evaluation
of \(p_0\), for example
$\nabla_{\mathbf{x}} p_0\,\frac{\partial \mathbf{X}_{\psi}}{\partial\psi}
+
\nabla_{\boldsymbol{\omega}} p_0\,\frac{\partial\boldsymbol{\omega}_{\psi}}{\partial\psi},$
analogous to the pixel-filter derivative discussed by Yu
\etal~\cite{10.1145/3550454.3555500}. For a discontinuous box
reconstruction filter, this also introduces the associated
pixel-boundary contribution. By keeping the sampled camera ray fixed,
the hemispherical parameterization evaluates the reconstruction filter
at a fixed sensor sample. The geometry dependence is instead represented
by the moving camera-ray intersection
\(\mathbf{x}_1=r_{\psi}(\mathbf{x}_c,\boldsymbol{\phi})\).

\section{Problem Setup and Training}
\label{sec:training}

The preceding formulation is written for recursive light transport and therefore admits multi-bounce path contributions. In this work, however, we instantiate its direct-illumination specialization for sparse-view geometry reconstruction. Given posed reference images, calibrated camera intrinsics and extrinsics, and known static point lights, we optimize the beta-surfel parameters $\psi_k$ defined in Eq.~\eqref{eq:surfel_parameters}.

The optimized objective combines image reconstruction with geometric regularization:
\begin{equation}
\TrainLoss =
\mathcal{J}_{\mathrm{rgb}}
+
\lambda_d \mathcal{L}_d
+
\lambda_n \mathcal{L}_n
+
\lambda_{\mathrm{weak}}\mathcal{L}_{\mathrm{\alpha}}.
\label{eq:training_loss}
\end{equation}
where $\mathcal{J}_{\mathrm{rgb}}$ is the image reconstruction loss across all RGB channels, $\mathcal{L}_d$ is the depth-distortion regularizer, $\mathcal{L}_n$ is the normal-consistency regularizer adopted from 2DGS~\cite{huang20242d}, and $\mathcal{L}_{\mathrm{\alpha}}$ is a weak visibility-weighted opacity regularizer.

\subsubsection{Weak Opacity Regularizer}
A black background introduces an opacity/radi\-ance ambiguity where a partially transparent surfel with high reflected radiance can produce a similar image contribution to a more opaque surfel with lower reflected radiance. We therefore introduce a weak visibility-weighted regularizer that encourages opacity only for surfels that contribute to observed training pixels:
\begin{equation}
\mathcal{L}_{\mathrm{\alpha}}
=
\sum_{p \in \mathcal{P}}
\sum_{i \in \mathcal{H}(p)}
w_{p,i}
\left(1-\eta_i\right)^2,
\label{eq:opacity_prior}
\end{equation}
where $\mathcal{P}$ denotes the set of training pixel rays and $\mathcal{H}(p)$ is the front-to-back ordered list of surfels intersected by ray $p$. The visibility weight is
\begin{equation}
w_{p,i}
=
\tau_{p,i}\,\alpha_{p,i},
\label{eq:opacity_prior_weight}
\end{equation}
where $\tau_{p,i}$ is the accumulated transmittance before surfel $i$ and $\alpha_{p,i}$ is its effective opacity at the ray--surfel intersection. We stop gradients through $w_{p,i}$ when evaluating this prior. The regularizer therefore cannot be reduced by changing the ray composition and instead directly encourages the opacity parameter of contributing surfels. The weight $\lambda_{\mathrm{weak}}$ in Eq.~\eqref{eq:training_loss} is kept small, so the prior is intended to regularize the opacity/radiance ambiguity while retaining a small relative weight.

\begin{figure*}[t]
\centering
\setlength{\tabcolsep}{0.5pt}
\begin{tabular}{ccccc}
Teapot & LEGO & Horse & Dragon & Plant \\
\includegraphics[width=0.195\textwidth]{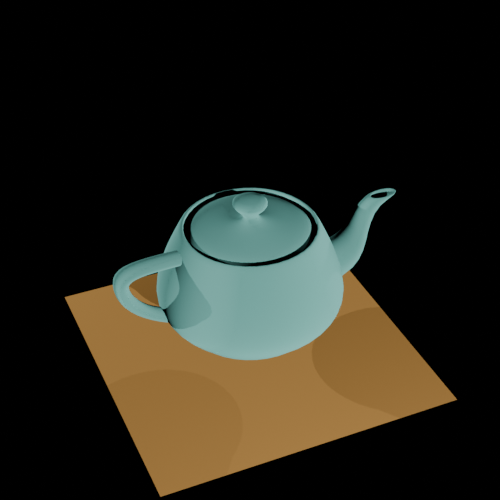} &
\includegraphics[width=0.195\textwidth]{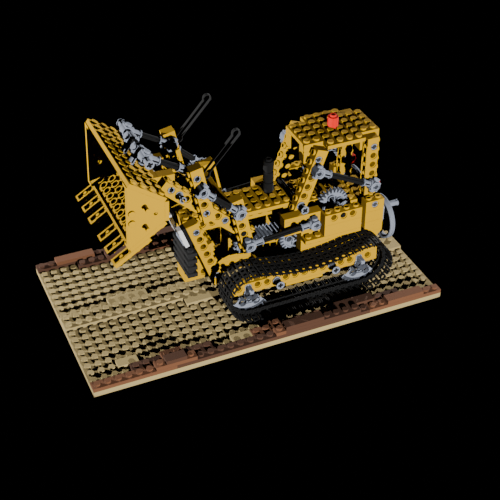} &
\includegraphics[width=0.195\textwidth]{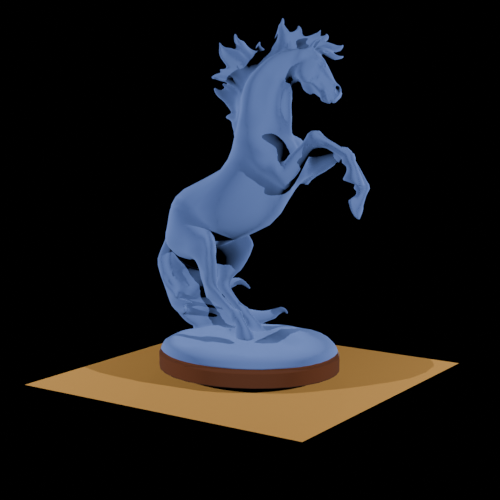} &
\includegraphics[width=0.195\textwidth]{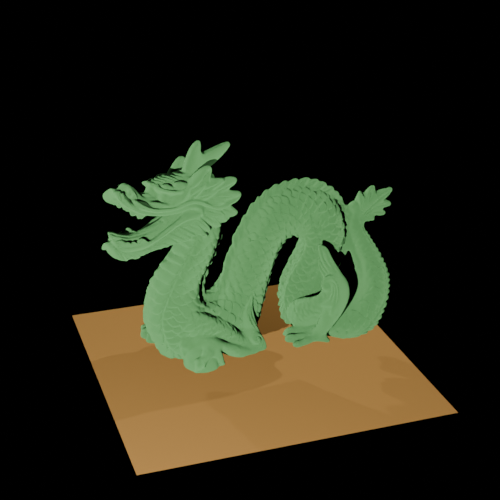} &
\includegraphics[width=0.195\textwidth]{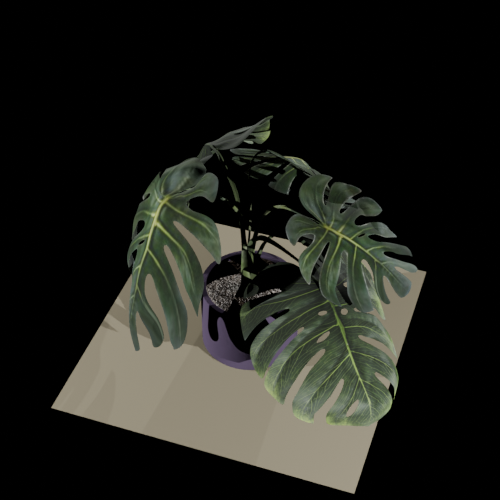}
\end{tabular}
\caption{
Overview of the synthetic sparse-view object reconstruction dataset.
Each scene is reconstructed from ten posed input views rendered with known static point-light illumination and Lambertian materials.
For visual clarity, we show one representative training image per scene.
}
\label{fig:dataset_overview}
\end{figure*}

% ---------------------------------------------------------------------
% Quantitative table
% ---------------------------------------------------------------------

\subsubsection{Densification and Pruning}
We adapt the number of surfels during optimization using clone-only densification based on position gradient and scale-based pruning. A surfel that is incorrectly positioned can reduce its image-space and light-transport contribution by shrinking its footprint. We therefore remove surfels whose scale falls below a fixed minimum threshold, as these primitives have become effectively unsupported by the reconstruction objective.

We do not use opacity pruning or periodic opacity resets commonly used by GS methods \cite{kerbl20233d}. Given a background color in the renderer, a surfel contribution is jointly controlled by opacity and reflected radiance, so reducing opacity can mimic changes in albedo or incident illumination. In preliminary experiments, opacity resets and opacity-based pruning removed useful surfels or encouraged low-opacity configurations that destabilized the normal consistency regularizer. We therefore avoid resetting opacity and remove unsupported surfels only through scale-based pruning.

\subsection{Dataset, Baselines, and Evaluation}
\label{sec:experimental_setup}
We evaluate on five synthetic object scenes: \emph{Teapot} \cite{Turk1994Zippered}, \emph{LEGO} \cite{mildenhall2020nerf}, \emph{Dragon} \cite{curless1996volumetric}, \emph{Horse}, and \emph{Plant}. The \emph{Horse} and \emph{Plant} assets were obtained from Blendkit under their respective licenses. We release only rendered data, calibration, and scene-generation code, not the original meshes or textures. Figure~\ref{fig:dataset_overview} shows one representative training image per scene. Each object is reconstructed from ten posed images at $500\times500$ resolution. The camera intrinsics and extrinsics, together with the positions and intensities of three to four static near-field point lights, are known to our method.

The scenes contain isolated Lambertian objects on black backgrounds. Although the geometry is relatively simple, ten views leave substantial portions of the surface weakly observed or unobserved. This setting therefore retains ambiguity between geometry, reflectance, and visibility while allowing controlled evaluation of known-light supervision.

Reference images are rendered in Blender Cycles using the same direct point-light illumination model assumed during optimization, Lambertian materials, and no indirect illumination. Blender serves as an independent renderer and avoids coupling the dataset-generation implementation to our optimization renderer.

\paragraph{Initialization.}
For all scenes, we initialize every point-based method from the same set of 25 white primitives positioned on a uniform grid within the scene
bounding volume. The primitives share identical initial positions, white appearance, scale \(0.1\), opacity \(0.5\), and vertical
orientation. For our beta surfels, we additionally initialize the shape parameter as \(\beta_k=0\).

\paragraph{Baselines.}
Our primary comparisons are against the point-based methods 2D\-GS~\cite{huang20242d}, SuGaR~\cite{guedon2023sugar}, RadiosityGS~\cite{jiang2025differentiable}, and GOF~\cite{yu2024gaussianopacityfieldsefficient}. We additionally include NeuS~\cite{wang2023neuslearningneuralimplicit} and NeuS2~\cite{neus2}, which use implicit neural surface representations, and GeoSVR~\cite{li2025geosvr}, which uses a sparse-voxel representation, to contextualize the point-based results within the broader landscape of surface reconstruction methods.

For 2DGS, we report the standard 30K-iteration configuration from the original method. We additionally report a 7K baseline which is the same iteration count that SuGaR uses for 3DGS before applying their meshes extraction procedure. For GeoSVR, we use the official implementation without its monocular depth constraint. Monocular depth predictions were unreliable for our black background object images without foreground masks and introduced incorrect geometric constraints. No foreground masks or external monocular depth priors are provided to the evaluated methods.

\subsection{Implementation Details}
\label{sec:implementation}

All experiments use a custom physically based renderer implemented in modern C++ and accelerated with SYCL \cite{sycl2020}. We tested the same implementation on AMD and NVIDIA GPUs using the corresponding SYCL backends. Optimization requires approximately 5--8 GB of VRAM and uses Adam~\cite{kingma2017adam} through PyTorch~\cite{paszke2019pytorch}, with default optimizer settings except for parameter-specific learning rates.

We selected the regularization weights empirically in preliminary experiments and fixed them across all reported scenes avoiding per scene specific tuning:
$\lambda_d=100$,
$\lambda_n=0.005$, and
$\lambda_{\mathrm{weak}}=0.05$
in Eq.~\eqref{eq:training_loss}. Our optimization runs for 60K iterations.
We extract meshes by truncated signed distance function (TSDF) fusion of depth maps rendered from the optimized surfels at the input camera poses, following the depth-fusion procedure used by 2DGS~\cite{huang20242d}. Ground-truth geometry is not used during optimization or mesh extraction.

\paragraph{Evaluation.}
We report symmetric Chamfer distance between reconstructed and ground-truth meshes. Accuracy denotes reconstruction-to-ground-truth distance and penalizes spurious geometry, while completion denotes ground-truth-to-reconstruction distance and penalizes missing surface regions.

\begin{table*}[!t]
\centering
\small
\setlength{\tabcolsep}{4.5pt}

\caption{
Main geometry reconstruction results for sparse-view object reconstruction.
Entries report symmetric Chamfer distance (CD), scaled by $10^2$ for readability; lower is better.
The mean is computed across all five scenes.
The final column reports the average number of optimized primitives for point-based methods.
Within the point-based methods, the
\colorbox{bestred}{best},
\colorbox{secondorange}{second-best}, and
\colorbox{thirdyellow}{third-best}
results are highlighted.
}

\label{tab:main_chamfer_results}

\begin{tabular}{@{}lrrrrrrr@{}}
\toprule
Method & Teapot & LEGO & Horse & Dragon & Plant & Mean $\downarrow$ & Avg pts $\downarrow$ \\
\midrule
NeuS2~\cite{neus2}
& 95.850 & 112.344 & 90.242
& 101.955 & 93.212 & 98.721 & -- \\
NeuS~\cite{wang2023neuslearningneuralimplicit}
& 1.396 & 1.424 & 2.022
& 1.397 & 2.180 & 1.684 & -- \\
%NeuS~\cite{wang2023neuslearningneuralimplicit}
%&- & - & - & - & - & - & -- \\
%\midrule
GeoSVR~\cite{li2025geosvr} $\dagger$
& 3.667 & 1.446 & 2.054
& 3.255 & 10.296 & 4.144 & -- \\
\midrule

2DGS 7K~\cite{huang20242d}
& \secondbest{2.471} & \thirdbest{1.429} & \secondbest{1.702}
& \thirdbest{2.251} & \secondbest{2.891} & \secondbest{2.149}
& \secondbest{43.0k} \\

2DGS 30K~\cite{huang20242d}
& 53.570 & 30.563 & 39.075
& 53.149 & 15.917 & 38.455
& \thirdbest{51.0k} \\

SuGaR~\cite{guedon2023sugar}
& 5.837 & \best{1.406} & \thirdbest{1.993}
& 2.613 & 3.468 & \thirdbest{3.064}
& 421.4k \\

GOF~\cite{yu2024gaussianopacityfieldsefficient}
& \thirdbest{3.450} & 65.990 & 2.394
& \secondbest{2.074} & \thirdbest{3.186} & 15.419
& 72.4k \\

RadiosityGS~\cite{jiang2025differentiable}
& 31.621 & 24.400 & 27.327
& 28.256 & 22.867 & 26.894
& 107.4k \\

\midrule
Ours
& \best{1.387} & \secondbest{1.426} & \best{1.319}
& \best{1.297} & \best{2.252} & \best{1.536}
& \best{267} \\
\bottomrule
\end{tabular}
\parbox{\textwidth}{\footnotesize
$\dagger$ GeoSVR is evaluated without its monocular-depth prior;
see Section~\ref{sec:experimental_setup}.}
\end{table*}
% ---------------------------------------------------------------------
% Qualitative figure
% ---------------------------------------------------------------------
\begin{figure*}[!t]
\centering

\setlength{\tabcolsep}{0.0pt}
\begin{tabular}{@{}S Q Q Q Q Q Q@{}}
& GT & \textbf{Ours} & SuGaR & 2DGS-7K & 2DGS-30K & RadiosityGS \\
\midrule
\sceneLabel{Teapot} &
\qualimg{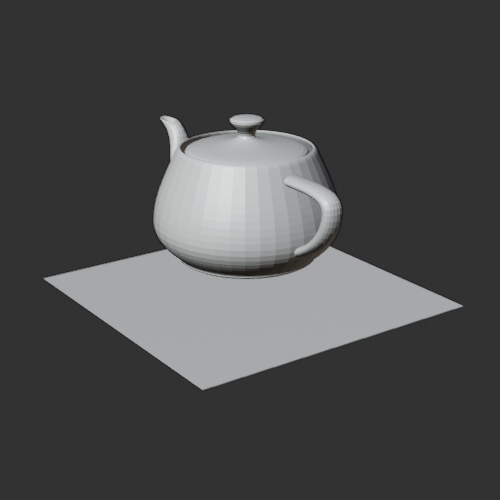} &
\qualimg{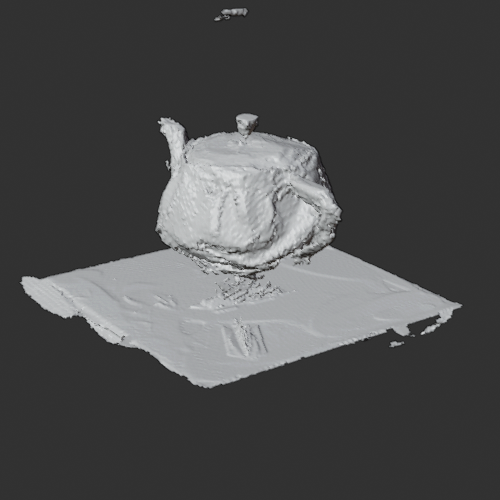} &
\qualimg{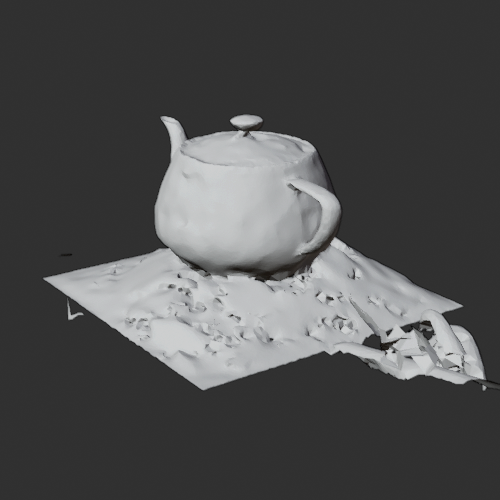} &
\qualimg{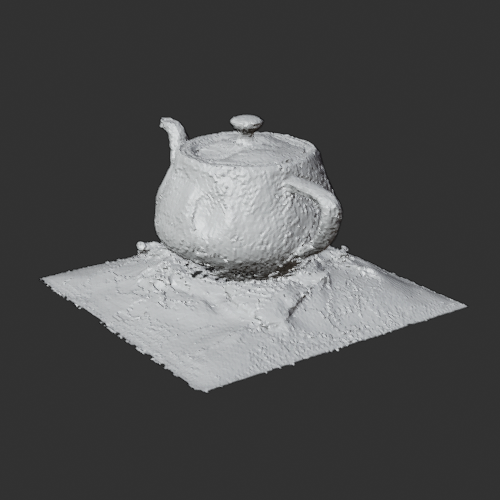} &
\qualimg{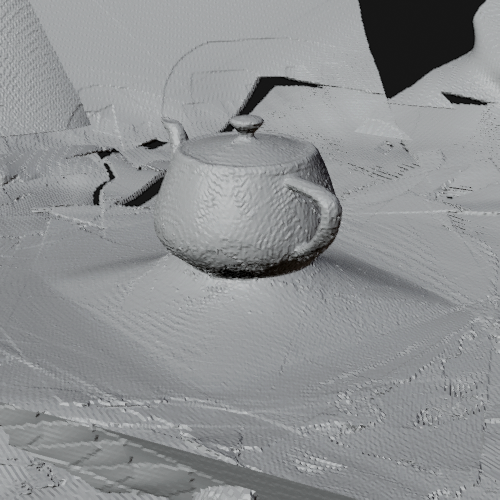} &
\qualimg{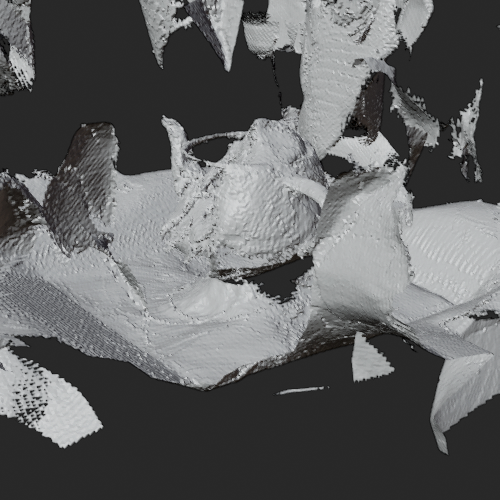}
\\
\sceneLabel{LEGO} &
\qualimg{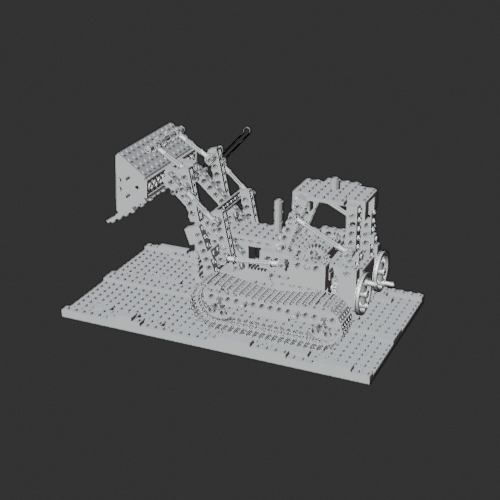} &
\qualimg{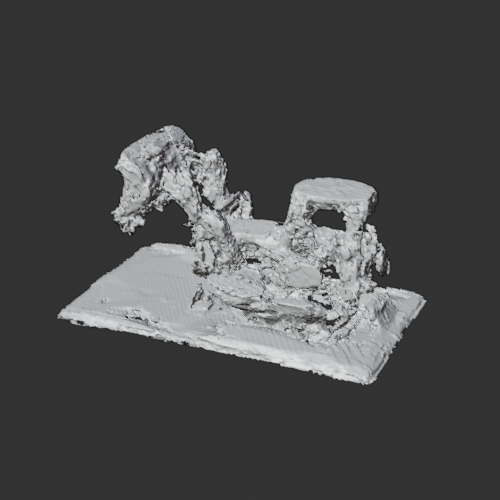} &
\qualimg{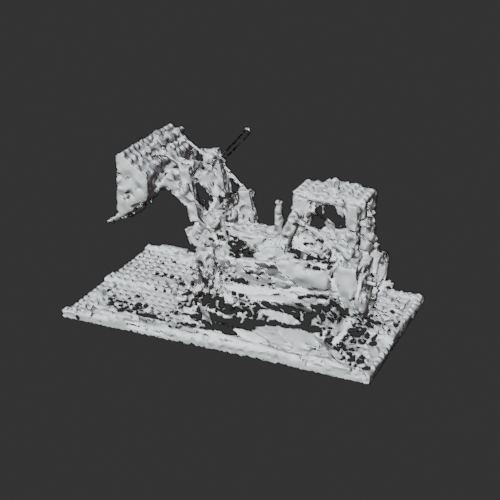} &
\qualimg{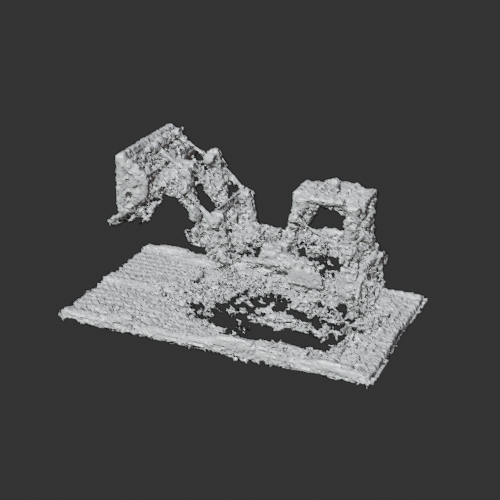} &
\qualimg{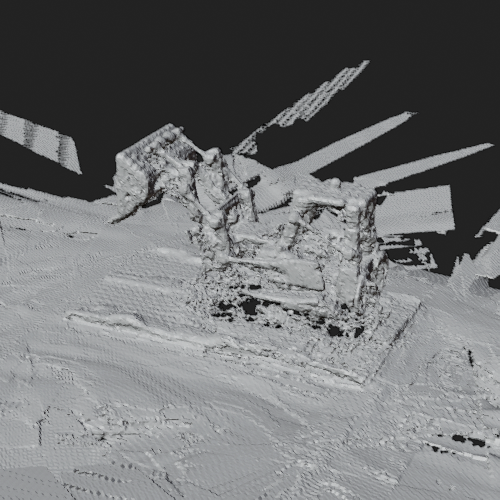} &
\qualimg{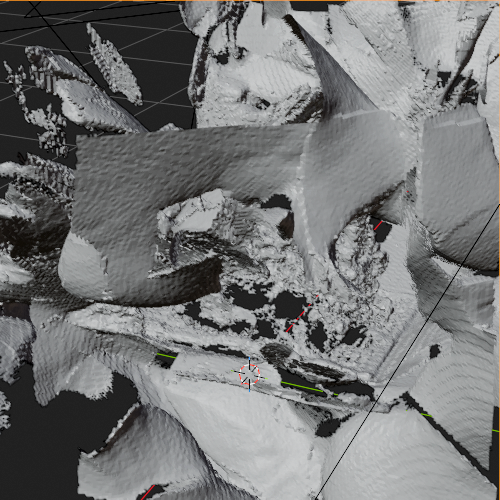}
\\
\sceneLabel{Horse} &
\qualimg{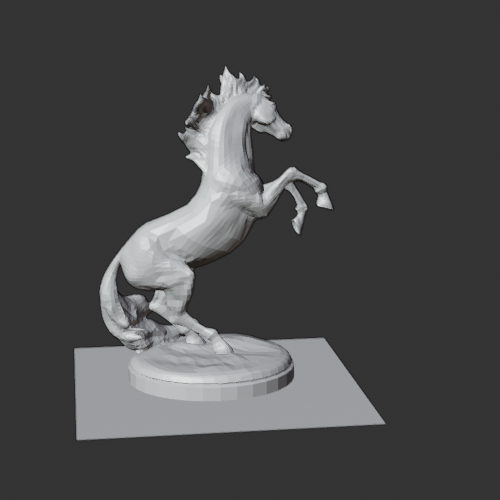} &
\qualimg{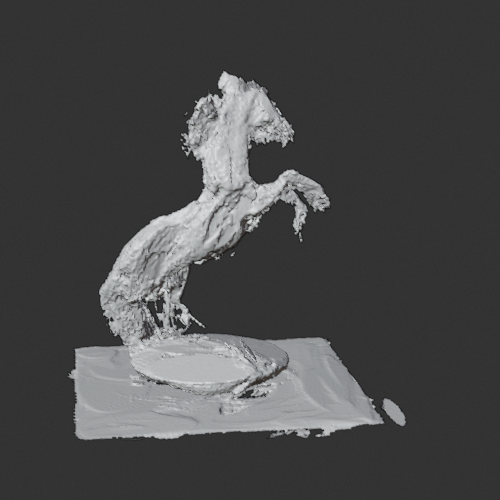} &
\qualimg{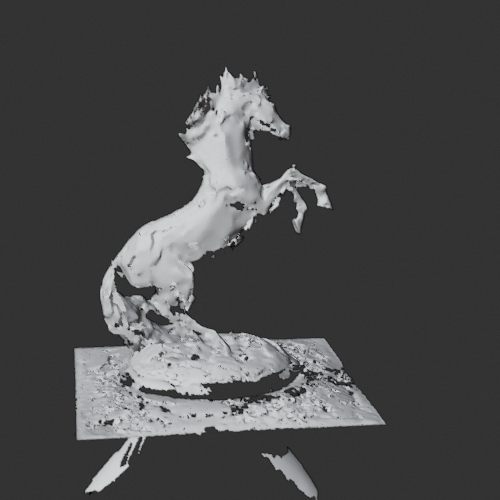} &
\qualimg{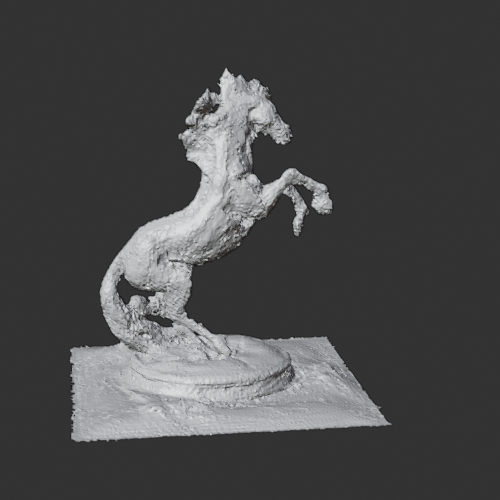} &
\qualimg{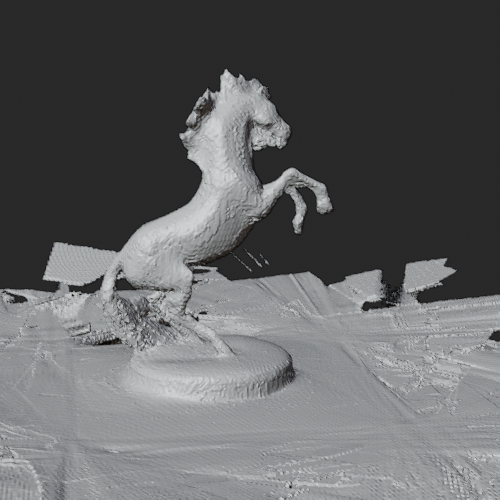} &
\qualimg{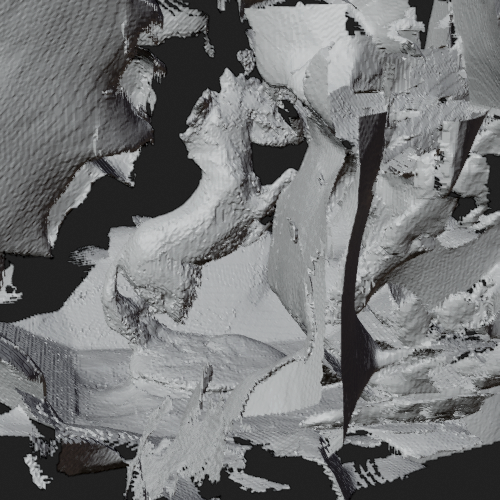}
\\
\sceneLabel{Dragon} &
\qualimg{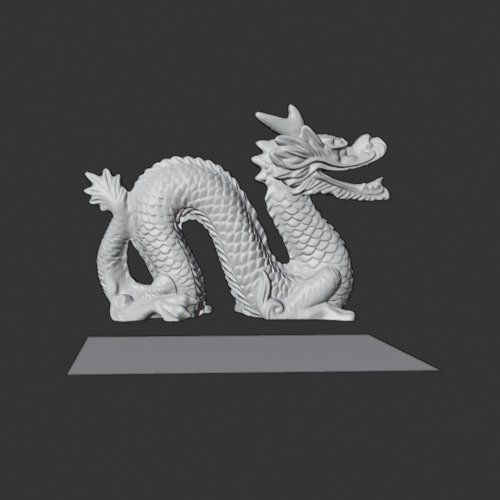} &
\qualimg{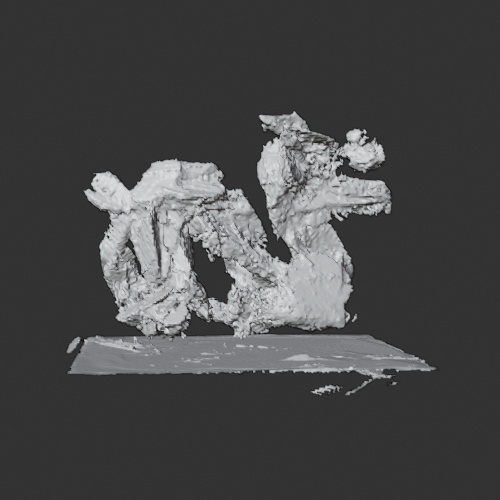} &
\qualimg{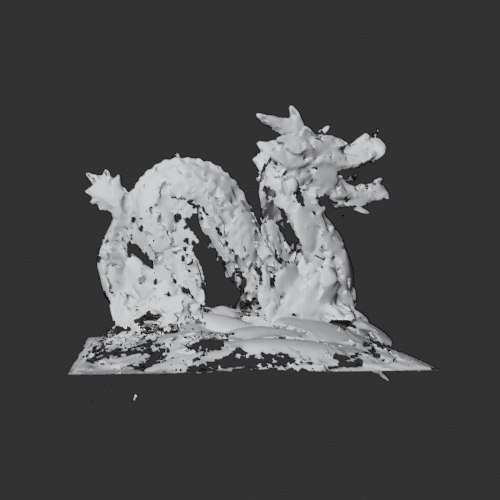} &
\qualimg{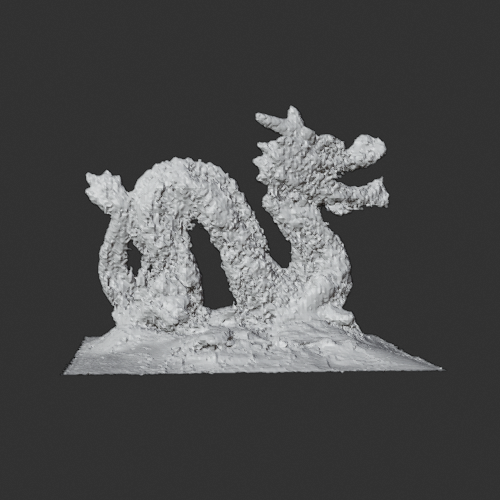} &
\qualimg{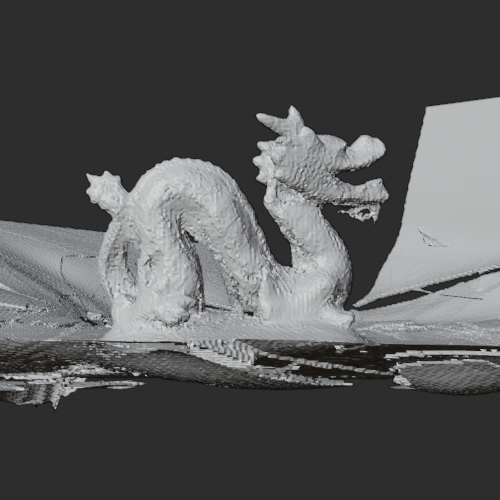} &
\qualimg{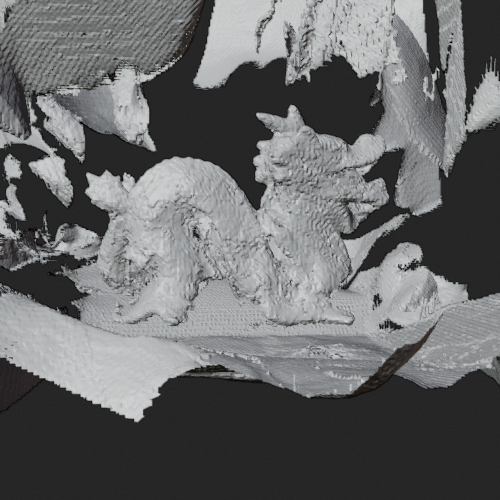}
\\
\sceneLabel{Plant} &
\qualimg{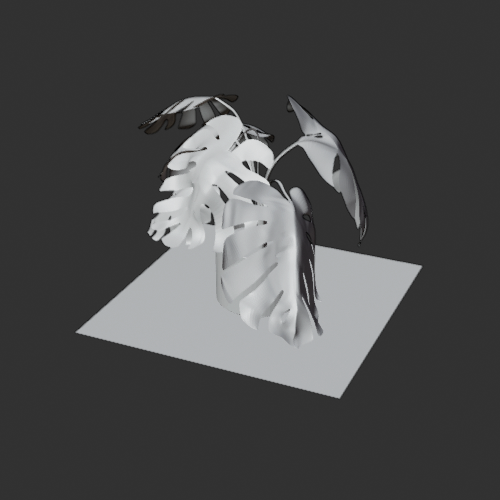} &
\qualimg{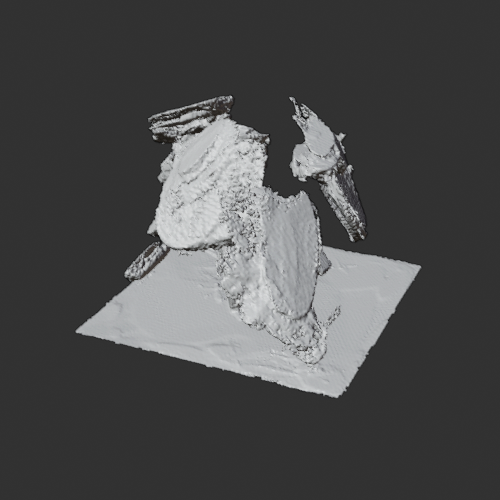} &
\qualimg{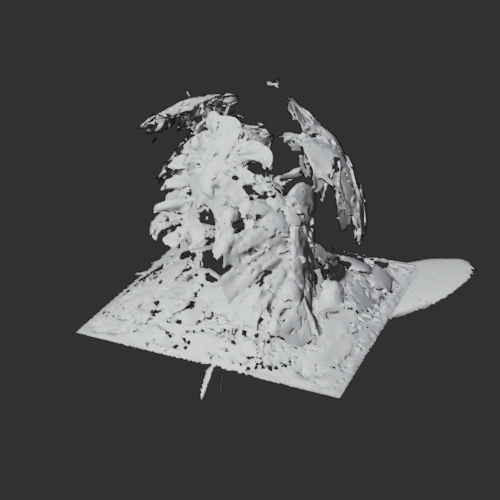} &
\qualimg{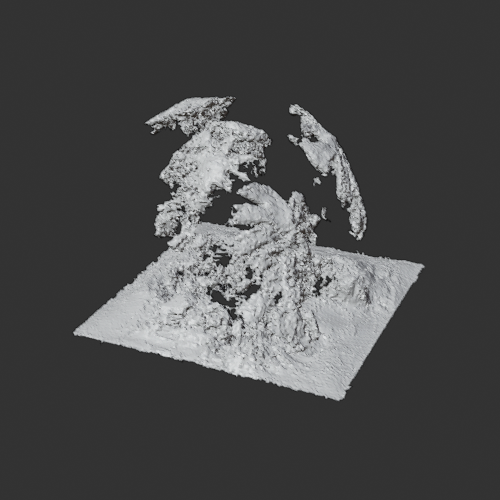} &
\qualimg{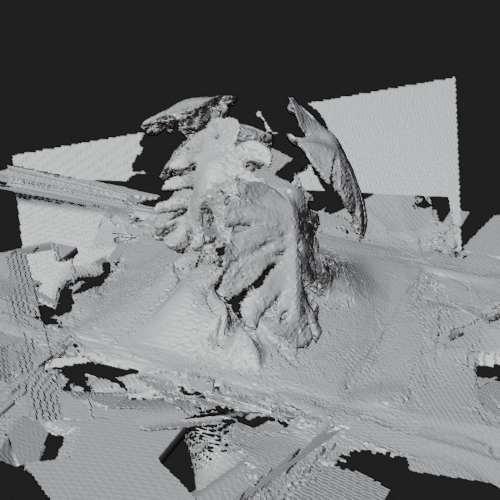} &
\qualimg{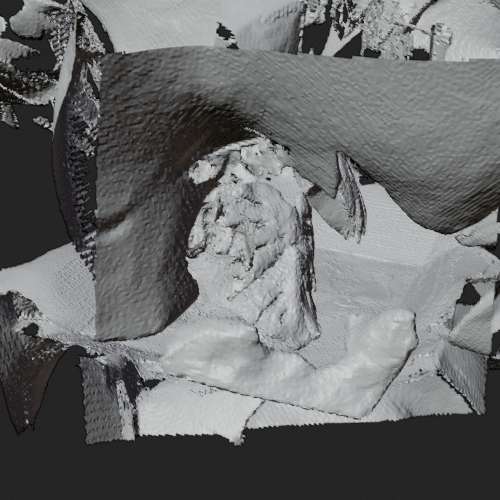}
\end{tabular}
\caption{
Qualitative reconstruction comparison across synthetic object scenes.
We compare the ground-truth geometry with our reconstruction and baseline methods.
Our method preserves the main object structure while using fewer optimized primitives.
}
\label{fig:qualitative_results}
\end{figure*}

\begin{table}[t]
\centering
\small
\setlength{\tabcolsep}{6pt}
\caption{
Unscaled mean directional Chamfer distance for the point-based methods across five scenes.
Accuracy penalizes spurious geometry, while completion penalizes missing regions.
}
\label{tab:mean_directional_chamfer}

\begin{tabular}{lcc}
\toprule
Method & Accuracy $\downarrow$ & Completion $\downarrow$ \\
\midrule
2DGS 7K~\cite{huang20242d}
& \secondbest{0.020} & \thirdbest{0.023} \\
2DGS 30K~\cite{huang20242d}
& 0.743 & 0.026 \\
SuGaR~\cite{guedon2023sugar}
& \thirdbest{0.037} & 0.024 \\
GOF \cite{yu2024gaussianopacityfieldsefficient}
& 0.295 & \best{0.014} \\
RadiosityGS~\cite{jiang2025differentiable}
& 0.512 & 0.026 \\

\midrule
Ours
& \best{0.017} & \best{0.014} \\
\bottomrule
\end{tabular}
\end{table}

\begin{table}[t]
\centering
\setlength{\tabcolsep}{8.5pt}

\caption{
Ablation study on the Horse scene using ten input views.
$\mathcal{L}_{d}$: depth distortion loss;
$\mathcal{L}_{n}$: normal consistency loss;
$\mathcal{L}_{\alpha}$: weak opacity prior.
CD denotes unscaled symmetric Chamfer distance; lower is better.
}
\label{tab:ablation_horse}

\begin{tabular}{lccccccc}
\toprule
Setting
& $\mathcal{L}_{d}$
& $\mathcal{L}_{n}$
& $\mathcal{L}_{\alpha}$
& Learn $\beta$
& CD $\downarrow$
& Acc. $\downarrow$
& Comp. $\downarrow$ \\
\midrule
Full
& \checkmark
& \checkmark
& \checkmark
& \checkmark
& \textbf{0.0132}
& \textbf{0.0138}
& \textbf{0.0125} \\

w/o $\mathcal{L}_{d}$
&
& \checkmark
& \checkmark
& \checkmark
& 0.0162
& 0.0171
& 0.0152 \\

w/o $\mathcal{L}_{n}$
& \checkmark
&
& \checkmark
& \checkmark
& 0.0185
& 0.0193
& 0.0178 \\

w/o $\mathcal{L}_{\alpha}$
& \checkmark
& \checkmark
&
& \checkmark
& 0.0241
& 0.0187
& 0.0295 \\

Fixed $\beta$
& \checkmark
& \checkmark
& \checkmark
&
& 0.0278
& 0.0245
& 0.0312 \\
\bottomrule
\end{tabular}
\end{table}

\section{Results}
\label{sec:results}

\subsubsection{Geometry Reconstruction}

Table~\ref{tab:main_chamfer_results} compares our method with implicit, sparse-voxel, and point-based reconstruction baselines. Our method achieves the lowest mean symmetric Chamfer distance and the best result on three scenes. Among the point-based methods, it performs best on four scenes, with SuGaR performing slightly better on \emph{LEGO}.

NeuS2 performs poorly without foreground masks or monocular-depth supervision. NeuS is the strongest non-point-based baseline, performing slightly better on \emph{LEGO} and \emph{Plant}, while our method performs better on the remaining scenes. GeoSVR is less accurate on average, particularly on \emph{Plant}, however, we omit its monocular-depth prior because its predictions were unreliable on these images.

We use the standard 30K-iteration 2DGS schedule as the primary baseline configuration. We additionally report the 7K checkpoint to illustrate an optimization behavior observed in our sparse-view setting. Although the 30K schedule further reduces the image-space objective, later iterations continue to introduce primitives near object boundaries that help explain the black background and reduce photometric residuals. These additional primitives do not improve the extracted surface and instead lead to substantially worse mesh accuracy.

The 7K checkpoint therefore achieves lower Chamfer error in our evaluation and indicates that continued densification and improved image fitting do not necessarily yield better geometry. 
Figure~\ref{fig:qualitative_results} illustrates this distinction for the 2DGS-30K and 7K results. In contrast, our method is optimized for 60K iterations without the same late-stage degradation in extracted geometry.

RadiosityGS is the closest physics-based point-based baseline. In our setting, it exhibits a related mismatch between image-space optimization and extracted mesh quality similar to 2DGS-30k. This suggests that, as with 2DGS, continued primitive growth can improve image fitting without improving the recovered surface. 
Nevertheless, the Dragon reconstruction in Figure~\ref{fig:qualitative_results} shows a notable benefit. RadiosityGS and our method are the only evaluated approaches that separate the object from the background plane, whereas the remaining methods fuse the two surfaces.

%Figure~\ref{fig:qualitative_results} compares the extracted meshes. Our method preserves the main object structures while producing fewer fragmented and spurious regions. While the rasterized approaches produces more smoother looking results. RadiosityGS is the closest physics-based point-rendering baseline, but produces substantially less accurate extracted geometry despite using a much denser representation. In our sparse-view evaluation, RadiosityGS does not achieve comparably accurate extracted geometry.

The 2DGS-30K reconstructions can appear smoother in some examples, but visual smoothness does not consistently correspond to geometric fidelity. The directional Chamfer results in Table~\ref{tab:mean_directional_chamfer} show that our improvement is not obtained merely by omitting uncertain regions: our method improves accuracy while maintaining competitive completion. Table~\ref{tab:main_chamfer_results} further shows that this improvement is achieved using only 267 surfels on average, compared with tens to hundreds of thousands of primitives for the point-based baselines. This combination of lower geometric error and a substantially smaller representation suggests a favorable scaling direction where increasing scene complexity need not require the rapid primitive growth observed for the evaluated splatting-based approaches.

These results indicate accurate recovery of the overall object geometry and major surface structures rather than fine-scale detail, making the method more suitable for applications such as coarse path planning, collision-aware navigation, and compact scene representation than high-fidelity digitization.

\paragraph{Ablations}
Table~\ref{tab:ablation_horse} evaluates the main components of the method on the Horse scene. We include depth distortion and normal regularization to show that they also benefit our optimization similar to 2DGS \cite{huang20242d}.
The weak visibility-weighted opacity regularizer has a larger impact on the result than depth and normal regularization and shows that for our simple scenes reducing the opacity/radiance ambiguity is beneficial. On Horse, fixing the beta profile similarly degrades reconstruction quality, indicating that the chosen opacity kernel is beneficial over a Gaussian-like shape only. This isolates the benefit of adapting the local opacity falloff during optimization.
\section{Conclusion}

We presented a geometry first point-based reconstruction method based on
compact opacity-bearing beta surfels and adjoint light transport with explicit transmittance. Across five synthetic objects reconstructed from ten posed views, the method achieves the lowest mean symmetric Chamfer distance among the evaluated baselines while using only 267 surfels on average. These results indicate that physically informed optimization can recover accurate coarse scale geometry without relying on dense primitive sets.

The present system is a deliberately constrained first instantiation and is not intended as the final form of the method. We evaluate isolated synthetic objects under known direct illumination and Lambertian reflectance, although the formulation admits recursive transport. Extending the system to indirect illumination, jointly estimated illumination, non-Lambertian reflectance, real captures, and more adaptive surfel refinement may provide stronger reconstruction cues. The current results therefore establish the viability of compact transport-optimized surfels while leaving substantial transport and
representational capacity unexplored.
% ---------------------------------------------------------------
% Bibliography
\bibliographystyle{splncs04}
\bibliography{main}

\newpage
\appendix

\section{Continuous Adjoint Formulation with Explicit Transmission}
\label{appendix_a}

This section provides the derivation of the opacity-explicit adjoint gradient used in the main paper. For completeness, we restate the rendering equation as
\begin{equation}
\label{eq:rendering}
L_o
=
\tau_{\psi}
\left(
L_e + \mathcal{T}_{\psi}L_o
\right),
\end{equation}
where $L_o$ is outgoing radiance, $L_e$ is emitted radiance, $\mathcal{T}_{\psi}$ is the light-transport operator, and $\tau_{\psi}$ is the beam-transmittance operator induced by the opacity-bearing scene representation. The image-space objective is written as
\begin{equation}
\mathcal{J}(L_o,\psi)
=
\left\langle
W_c,
J(L_o,\psi)
\right\rangle,
\end{equation}
where $W_c$ is the camera response and $J$ is the per-measurement loss.

From Eq.~\eqref{eq:rendering}, the corresponding state constraint is
\begin{equation}
\label{eq:state}
R(L_o,\psi)
=
-L_o
+
\tau_{\psi}\mathcal{T}_{\psi}L_o
+
\tau_{\psi}L_e.
\end{equation}
To obtain an unconstrained problem, we introduce a Lagrange multiplier $p$. The Lagrangian is
\begin{eqnarray}
\mathcal{L}(L_o,p,\psi)
&=&
\mathcal{J}(L_o,\psi)
+
\left\langle
p,
R(L_o,\psi)
\right\rangle
\nonumber\\
&=&
\left\langle
W_c,
J(L_o,\psi)
\right\rangle
+
\left\langle
p,
R(L_o,\psi)
\right\rangle,
\label{eq:lagrangian}
\end{eqnarray}
with
\begin{equation}
\left\langle f,g\right\rangle
=
\int_{A(\psi)}
\int_{4\pi}
f(x,\omega)g(x,\omega)
\,\ud\omega\,\ud A.
\end{equation}

Stationarity of the Lagrangian with respect to radiance,
$\partial\mathcal{L}/\partial L_o=0$, together with Eq.~\eqref{eq:state}, gives the adjoint equation
\begin{equation}
\label{eq:adjoint}
\frac{\partial J}{\partial L_o}W_c
=
-\left(
\frac{\partial R}{\partial L_o}
\right)^{\ast}p
=
-\left(
\tau_{\psi}\mathcal{T}_{\psi}-\mathcal{I}
\right)^{\ast}p
=
\left(
\mathcal{I}-\mathcal{T}_{\psi}^{\ast}\tau_{\psi}
\right)p,
\end{equation}
where $\ast$ denotes the adjoint and $\mathcal{I}$ is the identity operator. We treat beam transmittance as self-adjoint, $\tau_{\psi}^{\ast}=\tau_{\psi}$. A more detailed account of the transport operator and its adjoint is provided by Christensen~\cite{christensen2003adjoints}.

When the adjoint equation is satisfied, stationarity with respect to the scene parameters is an optimality condition, and $\partial\mathcal{L}/\partial\psi$ equals the gradient of the cost function~\cite{plessix2006review}. Before reparameterization, this derivative can be written as
\begin{equation}
\label{eq:optgrad}
\frac{\partial\mathcal{L}}{\partial\psi}
=
\frac{\partial\mathcal{J}}{\partial\psi}
+
\left\langle
p,
\frac{\partial R}{\partial\psi}
\right\rangle
+
\left\langle
p,
R
\left(
\nabla_A\!\cdot\!\frac{\partial x}{\partial\psi}
\right)
\right\rangle,
\end{equation}
where the final term accounts for the parameter dependence of the integration domain $A(\psi)$~\cite{sokolowski1992introduction}. In our implementation, surface integrals over individual surfels are reparameterized over fixed local coordinates. The corresponding parameter dependence is therefore represented through the differentiated integrand and the local surface-area Jacobian, rather than through an explicit moving-domain term.

\subsection{Adjoint Light Transport}

The rendering equation in Eq.~\eqref{eq:rendering} admits the Neumann-series expansion~\cite{kajiya1986rendering}
\begin{equation}
\label{eq:radiance}
L_o
=
\left(
\mathcal{I}-\tau_{\psi}\mathcal{T}_{\psi}
\right)^{-1}
\tau_{\psi}L_e
=
\sum_{n=0}^{\infty}
\left(
\tau_{\psi}\mathcal{T}_{\psi}
\right)^n
\tau_{\psi}L_e,
\end{equation}
where each application of the transport operator corresponds to one light-transport bounce. As observed by Stam~\cite{stam2020computing}, the adjoint equation has the same structure but uses the adjoint transport operator $\mathcal{T}_{\psi}^{\ast}$. The adjoint operator propagates an incident quantity, such as radiance or importance, rather than an outgoing quantity~\cite{christensen2003adjoints}. The adjoint variable is therefore
\begin{equation}
\label{eq:p}
p
=
\sum_{n=0}^{\infty}
\left(
\mathcal{T}_{\psi}^{\ast}\tau_{\psi}
\right)^n
\frac{\partial J}{\partial L_o}W_c
=
\sum_{n=0}^{\infty}
\left(
\mathcal{T}_{\psi}^{\ast}\tau_{\psi}
\right)^n
\left(
L_o-L_{\mathrm{ref}}
\right)W_c.
\end{equation}
Importantly, tracing $p$ depends neither on the number of optimized parameters nor on their individual derivatives.

Because eye-path tracing is the natural implementation strategy in our differentiable renderer, we express the gradient using derivatives of $\mathcal{T}_{\psi}^{\ast}$ rather than derivatives of $\mathcal{T}_{\psi}$. Substituting Eq.~\eqref{eq:state} into Eq.~\eqref{eq:optgrad}, taking the partial derivative with respect to $\psi$ while holding the radiance state $L_o$ fixed, and applying the defining property of adjoint operators gives
\begin{equation}
\frac{\partial\mathcal{L}}{\partial\psi}
=
\frac{\partial\mathcal{J}}{\partial\psi}
+
\left\langle
p,
\frac{\partial}{\partial\psi}
\left(
-L_o
+
\tau_{\psi}\mathcal{T}_{\psi}L_o
+
\tau_{\psi}L_e
\right)
\right\rangle.
\end{equation}
This simplifies to
\begin{equation}
\label{eq:grad}
\frac{\partial\mathcal{L}}{\partial\psi}
=
\frac{\partial\mathcal{J}}{\partial\psi}
+
\left\langle
\frac{\partial
\left(
\mathcal{T}_{\psi}^{\ast}\tau_{\psi}
\right)}
{\partial\psi}
p,
L_o
\right\rangle
+
\left\langle
p,
\frac{\partial\tau_{\psi}}{\partial\psi}L_e
\right\rangle,
\end{equation}
where the light sources are assumed not to be parameterized. The first inner-product term differentiates transported adjoint importance through scattering and transmission, while the second differentiates the transmission of emitted radiance. The surface-position integral in the source term for $p$ collapses through the Dirac delta contained in $W_c$ [Eq.~\eqref{eq:p}], allowing the gradient to be evaluated by differentiable path tracing from the camera.

\end{document}